\documentclass[runningheads]{llncs}
\usepackage{graphicx}
\usepackage{todonotes}
\usepackage{amsmath}
\usepackage{enumitem}
\usepackage{caption}
\usepackage{subcaption}
\usepackage[utf8]{inputenc}
\usepackage{hyperref}

\newlist{questions}{enumerate}{2}
\setlist[questions,1]{label=RQ\arabic*.,ref=RQ\arabic*}
\setlist[questions,2]{label=(\alph*),ref=\thequestionsi(\alph*)}

\begin{document}
%
\title{Prescriptive Process Monitoring Under Resource Constraints: A Causal Inference Approach\thanks{Supported by the European Research Council (PIX Project).}}

\titlerunning{Prescriptive Monitoring Under Resource Constraints}
%
\author{Mahmoud Shoush\orcidID{0000-0002-7423-9909} \and
Marlon Dumas \orcidID{0000-0002-9247-7476} }
\authorrunning{Mahmoud Shoush and Marlon Dumas}
%
\institute{University of Tartu, Tartu, Estonia\\
\email{\{mahmoud.shoush, marlon.dumas\}@ut.ee}
}
\maketitle              
%

\vspace{-4mm}
\begin{abstract}
Prescriptive process monitoring is a family of techniques to optimize the performance of a business process by triggering interventions at runtime. Existing prescriptive process monitoring techniques assume that the number of interventions that may be triggered is unbounded. In practice, though, interventions consume resources with finite capacity. For example, in a loan origination process, an intervention may consist of preparing an alternative loan offer to increase the applicant's chances of taking a loan. This intervention requires time from a credit officer. Thus, it is not possible to trigger this intervention in all cases. This paper proposes a prescriptive monitoring technique that triggers interventions to optimize a cost function under fixed resource constraints. The technique relies on predictive modeling to identify cases that are likely to lead to a negative outcome, in combination with causal inference to estimate the effect of an intervention on a case's outcome. These estimates are used to allocate resources to interventions to maximize a cost function. A preliminary evaluation suggests that the approach produces a higher net gain than a purely predictive (non-causal) baseline.

\end{abstract}
\vspace*{-9mm}

\section{Introduction}


\textit{Prescriptive Process Monitoring (PrPM)}~\cite{fahrenkrog2019fire,metzger2020triggering} is a set of techniques to recommend or to trigger actions (herein called \emph{interventions}) during the execution of a process in order to optimize its performance.
PrPM techniques use business process execution logs (a.k.a.\ \emph{event logs}) to predict negative outcomes that affect the performance of the process and use these predictions to determine if and when to trigger interventions to prevent or mitigate such negative outcomes. 

Several PrPM techniques have been proposed~\cite{fahrenkrog2019fire,metzger2020triggering,bozorgi2021prescriptive}. These techniques, however, assume that it is possible to trigger any number of interventions at any point in time. In practice, each intervention requires some resources (e.g., time from an employee), and those resources have a limited capacity. 
For example, in a loan origination process, an intervention could be to provide an alternative loan offer to increase the applicant’s likelihood of taking a loan. This intervention can only be triggered if a loan officer is available to perform it.

In this setting, we address the question of whether or not to trigger an intervention during the execution of an instance of a business process (herein a \textit{case}) to optimize a gain function that considers the cost of the case ending in a negative outcome and the cost of the intervention. We tackle this question in the context where each intervention requires locking a resource for given \textit{treatment duration} and where the number of available resources is bounded. 

To address this question, we use a predictive model to estimate the probability of a negative case outcome and a causal inference approach to estimate the effect of triggering an intervention on the probability of a negative case outcome. Based on these estimates, we estimate the gain of triggering an intervention for each case. We use this estimate to decide which cases should be treated given the available resources.
We report on  evaluating a real-life event log to compare the proposed approach with a baseline that relies only on predictive models.

The paper is structured as follows. Sect.~\ref{sec:bgrw} presents background concepts and related work. Sect.~\ref{sec:approach} explains the approach while Sect.~\ref{sec:evaluation} discusses the empirical evaluation. Finally, Sect.~\ref{sec:conclusion} summarizes this paper and future work directions.

\vspace{-3mm}
\section{Background and Related Work} \label{sec:bgrw}
\vspace{-1mm}



\subsection{Predictive Process Monitoring}
\vspace{-1mm}
PrPM techniques are closely related to techniques for estimating the probability of negative case outcomes, also known as outcome-oriented Predictive Process Monitoring (PPM) techniques~\cite{teinemaa2019outcome}. The input of an outcome-oriented PPM technique is an \emph{event log} representing the execution of a business process. 
An extract of a loan handling process is shown in Fig.~\ref{fig:runningExample}. This log consists of two traces. Each trace consists of a sequence of \emph{events}. An event describes the execution of one activity instance. An event contains three attributes: a case identifier ($c_{id}$), an activity label (\emph{activity}), and a \emph{timestamp}. 
Other event attributes may exist, like who does the activity (the \emph{resource}). Additional attributes may be of one of two types: \emph{case attributes} or \emph{event attributes}. Case attributes are attributes whose values do not change within a case, while the value of an event attribute changes. For example, in Fig.~\ref{fig:runningExample}, the log contains two case attributes (\emph{age} and \emph{gender}) and one event attribute (\emph{resource}).
\vspace{-1mm}
\begin{figure*}[!th]
\vspace*{-8mm}

	\begin{center}
	\scalebox{0.7}{\parbox{\linewidth}{%
		\begin{align*}
             trace_1 &  = [(1, submitAnApplication, 12:00PM,  (resource, emp_1), (age, 25), (gender, male), \\ & \hspace{0.5cm} ...,
            (1, callClients,  02:00PM, (resource, emp_2), (age, 25), (gender, male))]\\
            trace_2 &  = [(2, makeAnOffer, 10:00AM, (resource, emp_3), (age, 30), (gender, female)), ...,\\
            & \hspace{0.5cm}(2, verifyDocuments, 02:00PM, , (resource, emp_4), (age, 30), (gender, female))]
        \end{align*}}}
		\label{fig:runningExample}
		\scalebox{0.7}{\parbox{\linewidth}{%
		\caption{Extract of a loan application process.}
		\label{fig:runningExample}}}
	\end{center}
	\vspace*{-10mm}

\end{figure*}

Outcome-oriented PPM methods predict the outcome of an ongoing case, given its (incomplete) trace. In a typical binary PPM method, the outcome of a case may be positive (e.g., a client accepted the loan offer) or negative (the client did not accept the offer). Accordingly, a precondition for applying a PPM method is to notion case outcomes and historical data about case outcomes. In the above example, this means that for each trace, we need to know whether or not the customer accepted the loan offer. An event log in which each trace is labeled with a case outcome is called a \emph{labeled event log}.

PPM methods typically distinguish between an offline training phase and an online prediction phase. Based on historical (completed) cases, a predictive model (specifically a classification model) is trained in the offline phase. This model is then used during the online phase to make predictions based on incomplete traces. A typical approach to train models for PPM is to extract all or a subset of the prefixes with length $k$ of the labeled trace in an event log and associate the full trace’s label to every prefix extracted from the trace. A dataset of this form is called a \emph{labeled prefix log}. A labeled prefix log is a set of prefixes of traces, each one with an associated case outcome (positive or negative).



\vspace{-2mm}

\begin{figure*}[htb]
\vspace*{-7mm}

	\begin{center}
	\scalebox{0.7}{\parbox{\linewidth}{%
		\begin{align*}
            vector_1  & = [((age, 25), (gender\_male, 1), (gender\_female, 0)), \\ 
            & \hspace{0.5cm}((res\_emp_1, 1), (res\_emp_2, 0), (res\_emp_3, 0), (res\_emp_4, 0)), \\ &
            \hspace{0.5cm}((A\_submit\_an\_application, 1), ((A\_communicate\_clients, 0), \\ &
            \hspace{0.5cm}((A\_make\_an\_offer, 0), ((A\_verify\_documents, 0)), (sum\_time, 0)]
        \end{align*}}}
		\scalebox{0.8}{\parbox{\linewidth}{%
		\caption{Aggregate encoding for  $trace_1$ with $k=1$.}
		\label{fig:runningExample2}}}
	\end{center}
	\vspace*{-14mm}

\end{figure*}
We use the labeled prefix log to train a machine learning algorithm to build a predictive monitoring model. However, we need first to encode the prefixes in the prefix log of each trace as so-called \emph{feature vectors} (herein called \emph{trace encoders}). Teinemaa et al. \cite{teinemaa2018temporal} propose and evaluate several types of trace encoders and find that \emph{aggregation encoder} consistently yields models with high accuracy.

An aggregate encoder is a function that maps each prefix of a trace to a feature vector. Simply, it encodes each case attribute as a feature (or one-hot encode categorical case attributes). For each numerical event attribute, use an aggregation method (e.g., sum) over the sequence of values taken by this attribute in the prefix. For every categorical event attribute, encode every possible value of that information as numerical features. This information refers to the number of times this value has appeared in the prefix. An example of applying aggregate encodings to  $trace_1$ with $k=1$ is shown in Fig.~\ref{fig:runningExample2}.

\vspace{-3mm}
\subsection{Prescriptive Process Monitoring}

Various PrPM methods have been proposed in prior work.   
Fahrenkrog et al.~\cite{fahrenkrog2019fire} introduce an approach to generate single or multiple alarms when the probability of a case leading to an undesired outcome is above a threshold (e.g., 70\%). Each alarm triggers an intervention, which reduces the probability of a negative outcome. Their method optimizes the threshold empirically w.r.t a gain function. 

Metzger et al.~\cite{metzger2020triggering} use ensemble methods to compute predictions and reliability estimates to trigger interventions. They introduce policy-based reinforcement learning to find and learn when to trigger proactive process adaptation. This work targets the problem of learning when to trigger an intervention rather than the question of whether or not to trigger an intervention. Both the technique of Metzger et al. and that of Fahrenkrog et al. work under the assumption that the number of interventions that may be triggered at a given point in time is unbounded. In contrast, in this paper, we consider resource constraints.


Weinzerl et al.~\cite{sven2020} propose a PrPM technique to recommend the next activity in each ongoing case of a process, to maximize a given performance measure. This previous study does not consider an explicit notion of intervention. Thus, it does not consider the cost of intervention nor the fact that an intervention may only be triggered if a resource is available to perform it.



\vspace{-4mm}

\subsection{Causal Inference}
\emph{Causal Inference (CI)}~\cite{xu2020causality} is a collection of techniques to discover and quantify cause-effect relations from data. Causal inference techniques have been used in a broad range of domains, including process mining.

In \cite{bozorgi2020process}, the authors introduce a technique to find guidance rules following Treatment $\to$ Outcome relation, which improves the business process by triggering an intervention when a condition folds. They generate rules at design time in the level of groups of cases that will be validated later by domain experts. More recently, in \cite{bozorgi2021prescriptive}, they address another target problem: reducing the cycle time of a process using interventions to maximize a net gain function. Both works \cite{bozorgi2020process} and \cite{bozorgi2021prescriptive} consider the estimation of the treatment effect. However, they assume that interventions with a positive impact occur immediately and do not examine the finite capacity of resources. 



Causal inference techniques are categorized into two main frameworks \cite{guo2020survey}: (1) \textit{Structural Causal Models} (SCMs), which consist of a causal graph and structural equations [1]. SCM focuses mainly on estimating the causal effects through a causal graph which a domain expert manually constructs. (2) \textit{Potential outcome frameworks} focus on learning the treatment effects for a given treatment-outcome set $(T, Y)$. Our work utilizes the latter, which focuses on automatic estimation methods rather than manually constructed graphs. 

We use potential outcome models to estimate the treatment effect hereafter called \emph{conditional average treatment effect (CATE)} from observational data. In particular, we use an \emph{orthogonal random forest (ORF)} algorithm that combines tree-based models \cite{athey2019generalized} and double machine learning \cite{chernozhukov2018double} in one generalized approach \cite{oprescu2019orthogonal}. It estimates the $CATE$ on an outcome $Y$ when applying a treatment $T$ to a given case with features $X$.




ORF requires input to be in the form of $input = \{(T_i, Y_i, W_i, X_i)\}_{i=1}^n$ for $n$ instances. For each instance $i$, $T_i$ is described by a binary variable $T \in \{0, 1\}$, where $T = 1$ refers to treatment is applied to a case and $T = 0$ that it is not.  $Y_i$ refers to the observed outcome. $W_i$ describes potential confounding properties, and $X_i$ is the information achieving heterogeneity.

\vspace{-4mm}

\section{Approach} \label{sec:approach}
\vspace{-1mm}
The primary objective of our approach is to determine whether or not to treat a given case and when an intervention takes place to maximize the total gain. To learn whether or not to treat, we build predictive and prescriptive models in the \emph{learning phase}. Then, the \emph{resource allocator} selects when to treat.


\begin{figure*}[!htb]
	\begin{center}
        \resizebox{0.8\textwidth}{!}{\includegraphics{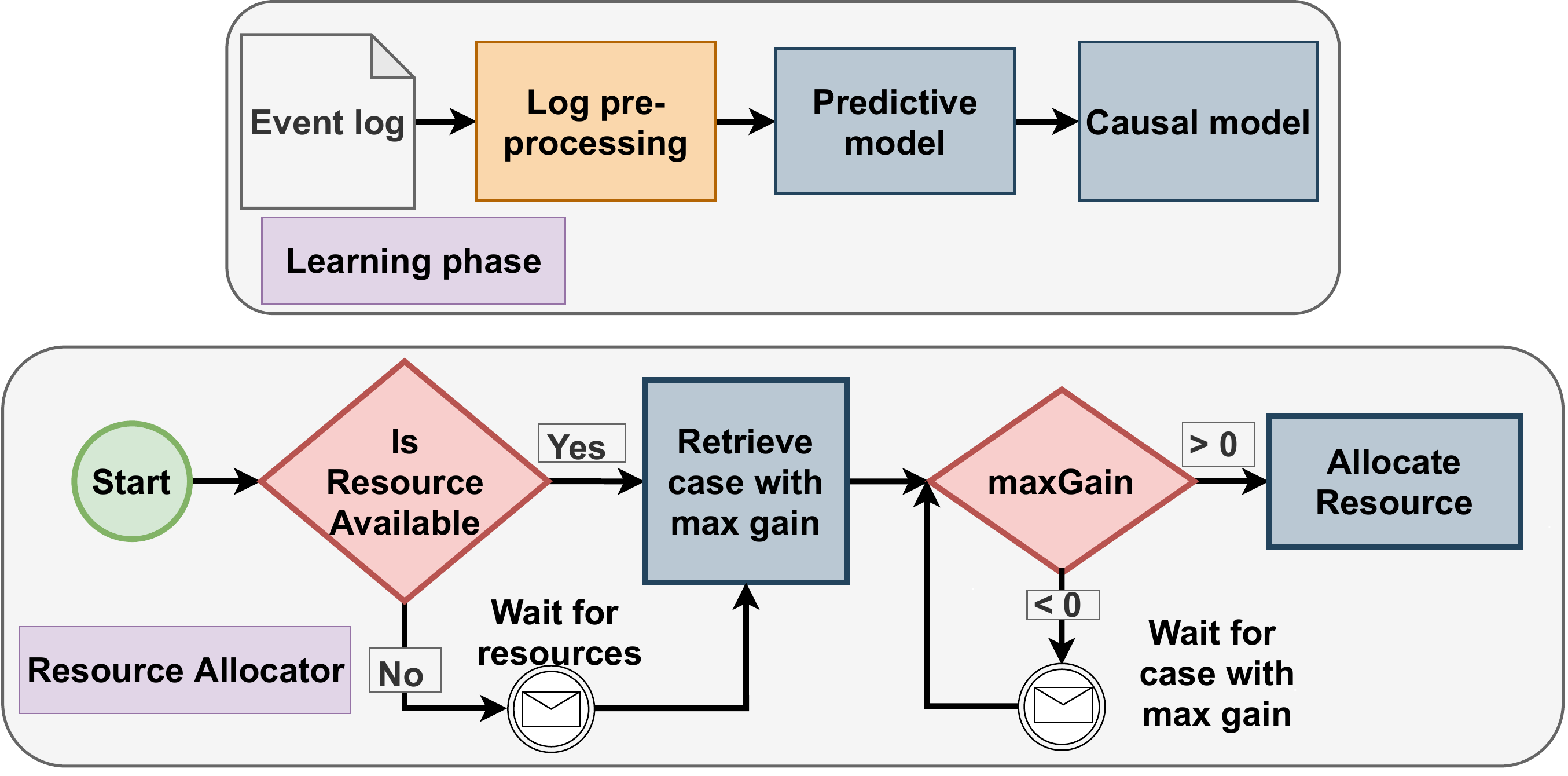}}
        \scalebox{0.8}{\parbox{\linewidth}{%
		\caption[]{Proposed approach}
		\label{fig:approach}}}
	\end{center}
	\vspace*{-5mm}

\end{figure*}

The approach consists of two phases, as shown in Fig.~\ref{fig:approach}. In the learning phase, we prepare the event log to build two machine learning models. The first one is a model that estimates the undesired case outcome probability. The second one is the causal model to estimate the impact of a given intervention on the outcome of a case. The predicted probability of the negative outcome and the estimated treatment effect are used to determine the net gain in the resource allocation phase. Below, we explain each step in Fig.~\ref{fig:approach}.

\vspace{-10mm}
\subsection{Log Preprocessing}
\vspace{-2mm}
Log preprocessing is an essential step in our approach that includes data cleaning, $k$-prefix extraction, prefix encoding, and identifying the outcome of cases and interventions that we might apply to reduce the probability of negative outcomes. For data cleaning, prefix extraction, and encoding, we follow the same approach proposed by Teinemaa et al. ~\cite{teinemaa2019outcome}. The setting of outcome and intervention is process-dependent which means we first need to understand the business process objective. Next, we analyze the log to find what interventions could affect the outcome of a given case by reducing the probability of negative outcomes.
\vspace{-1mm}

\vspace{-2mm}
\vspace{-2mm}
\subsection{Predictive Model}
\vspace{-2mm}
We build a predictive model to estimate the probability that cases will end with the undesired outcome. We use the estimated probabilities as a threshold $\tau$ that we optimize empirically to decide if we move forward to estimate the treatment effect and define gains or not. 
\vspace{-1mm}

\begin{figure*}[!htb]
\vspace{-7mm}
	\begin{center}
        \resizebox{0.8\textwidth}{!}{\includegraphics{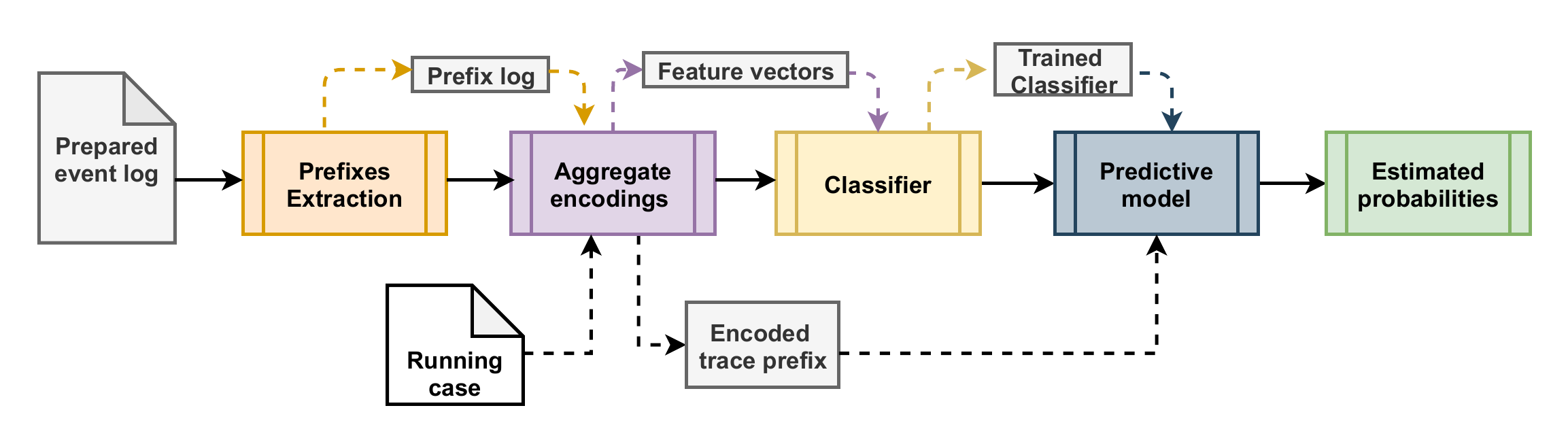}}
        \vspace{-4mm}
        \scalebox{0.8}{\parbox{\linewidth}{%
		\caption[]{Predictive model steps.}
		\label{fig:predictive_model}}}
	\end{center}
	\vspace*{-9mm}
\end{figure*}
\vspace{-2mm}

In order to build a predictive model, as shown in Fig \ref{fig:predictive_model}, first, we extract prefixes of length $k$ from every trace that results in a so-called \textit{prefix log}. This prefix extraction guarantees that our training log is similar to the testing log. For instance, If we have a complete trace containing seven events, we extract prefixes up to five events. Then we will have five incomplete traces starting with a trace containing only one event till a trace carrying five events. Next, in the aggregate encodings step, we encode each trace prefix into a fixed-size feature vector (see example in Fig \ref{fig:runningExample2}). Finally, we use the encoded log to train a machine learning method to estimate the probability of the undesired outcome. 

This article deals with an outcome-oriented PPM problem, a classification problem from a machine learning perspective. The output from training a classification technique is a predictive model to estimate the probability of the undesired outcome (i.e., $P_{uout}$) of running cases.
\vspace{-1mm}
\subsection{Causal Model}
\vspace{-1mm}


We use ORF to build a causal model to estimate the treatment effects or the $CATE$ of an intervention in a given case. An advantage of using ORF w.r.t.\ other causal models is that it handles well high-dimensional feature spaces. This is useful in our setting because event logs have many event attributes with categorical values, leading to feature explosion. 




To estimate CATE using ORF, the input needs to be in the form of $input = \{(T_i, Y_i, W_i, X_i)\}_{i=1}^n$ for $n$ instances. For each instance $i$, $T_i$ is the accepted treatment. $Y_i$ refers to the observed outcome. $T_i$ and $Y_i$ come from the preprocessing step (see Sect. 3.1), and they might be differ from one process to another based on the process objective. $W_i$ describes the potential confounding variables, and $X_i$ is the information achieving heterogeneity. In this work, we deal with an outcome-oriented loan application process it means the purpose is to increase the rate of successful loan applications via treating ongoing applications. We hypothesized that the intervention increases the number of successful applications, and we assume that the treatment is identified beforehand. $X$ and $W$ are obtained from the encoded log. $X$  is a feature vector that carries both case and event attributes, including the activity names, resources, and other extracted features,e.g., temporal attributes. We consider all these attributes to achieve the heterogeneity of the intervention effect. In this work, we assume that all log attributes $X$ are too possible confounders $W$. Nevertheless, $X$ and $W$ may not be the same variables where a domain expert can specify which features would be removed from $W$ if they do not improve the outcome.  

Next, and based on the above descriptions, we train an ORF to estimate the treatment effect. The output from training an ORF technique is a causal model used to estimate $CATE$ for running cases.


\vspace{-4mm}
\subsection{Resource Allocator}
\vspace{-1mm}

We trained two models in the learning phase: the predictive one to estimate the probability that a case will end with the undesired outcome $P_{uout}$ and the causal model to estimate the $CATE$ of utilizing an intervention in a given case. We use both models with the \emph{resource allocator} to decide whether or not to treat a given case and when the intervention takes place to maximize the \emph{total gain}.


Regularly triggering interventions in cases may come with gain; however, it comes at a cost. Therefore, to define the \emph{total gain}, we determine the costs with and without intervention if the predictive model gives a probability higher than a specific threshold $\tau$. Especially, suppose the intervention cost is relatively expensive as opposed to the advantage that it could afford. In that case, it becomes more critical to decide whether or not to treat a given case. 




A suitable threshold is not identified beforehand. One solution is to define and optimize the threshold empirically to obtain maximal gain instead of a random fixed value. The threshold is used to ensure that a given case has a high probability of ending with the undesired outcome, i.e., $P_{uout} > \tau$.

\vspace{-1mm}
\begin{definition}{\textit{\textbf{Cost with no intervention.}}}
	\textit{$cost(c_{id}, T_{i=0})$} The cost when $c_{id}$ ends with an undesired outcome without applying the intervention; therefore, $i=0$ is shown in equation \ref{eq:cnt}. The $P_{uout}$ is the estimated probability of the undesired outcome from the predictive model, and $c_{uout}$ is the cost of the undesired outcome.
	\vspace{-3mm}
	\begin{equation}
        cost(c_{id}, T_{i=0}) = P_{uout} * c_{uout}
        \label{eq:cnt}
    \end{equation}
\end{definition} 
\vspace{-2mm}

\vspace{-2mm}
\begin{definition}{\textit{\textbf{Cost with intervention.}}}
	\textit{$cost(c_{id}, T_{i=1})$} The cost when $c_{id}$ ends with an undesired outcome with applying the intervention; therefore, $i=1$ is shown in equation \ref{eq:ct}. The $CATE_1$ is the estimated causal effect of applying $T_{i=1}$ to $c_{id}$ resulting from the ORF model. $c_{T_1}$ is the cost of employing $T_{i=1}$ to $c_{id}$.  
	\vspace{-2mm}
	\begin{equation}
        cost(c_{id}, T_{i=1}) = (P_{uout} - CATE_1) * c_{uout} + c_{T_1}
        \label{eq:ct}
    \end{equation}
\end{definition} 
\vspace{-2mm}
Now, we have the costs with ($cost(c_{id}, T_{i=1})$) and without ($cost(c_{id}, T_{i=0})$) the intervention, the estimated probability ($P_{uout}$), and $CATE_1$ in our pocket. The next step is defining the $gain$ from applying $T_{i=1}$ to $c_{id}$ that enables the highest cost reduction based on equations \ref{eq:cnt} and \ref{eq:ct}, as shown in equation \ref{eq:gain}. The gain decides whether or not to treat $c_{id}$, which solves the first part of our problem.

\vspace{-2mm}

\begin{definition}{\textit{\textbf{Gain.}}}
	\textit{$gain(c_{id}, T_{i=1})$} 
	\vspace{-2mm}
	\begin{equation}
        gain(c_{id}, T_{i=1}) = cost(c_{id}, T_0)  - cost(c_{id}, T_{i=1})
        \label{eq:gain}
    \end{equation}
\end{definition}
\vspace{-2mm}

For example, suppose we have an event log with six cases (see table \ref{tab:gainex}), the $c_{uout} = 20$, and the $c_{T_1} = 1$. We have two situations where we do not calculate the costs with and without intervention and, therefore, the gain. The first one is presented with $c_{id} = C$ where the estimated probability is below a certain threshold, for instance, $\tau=0.5$. The other one is given with $c_{id} = F$, where there is no positive effect of applying intervention to the case; though, the $P_{uout} > \tau$. Other cases fulfill the conditions of having $P_{uout} > \tau$ and $CATE_1 > 0$.
\vspace{-7mm}
\begin{table}[hbpt]
\centering
\scalebox{0.8}{\parbox{\linewidth}{%
\caption{An example of defining gain.}
	\label{tab:gainex}}}
\resizebox{0.8\textwidth}{!}{
\begin{tabular}{lccccccc}
\hline
$c_{id}$ \hspace{0.5cm} & \multicolumn{1}{l}{$P_{uout}$} \hspace{0.5cm} & \multicolumn{1}{l}{$c_{uout}$} \hspace{0.5cm} & \multicolumn{1}{l}{$c_{T_1}$} \hspace{0.5cm} & \multicolumn{1}{l}{$CATE_1$} \hspace{0.5cm} & \multicolumn{1}{l}{$cost(c_{id}, T_0)$} \hspace{0.5cm} & \multicolumn{1}{l}{$cost(c_{id}, T_{i=1})$} \hspace{0.5cm} & \multicolumn{1}{l}{$gain(c_{id}, T_{i=1})$} \\ \hline
A        & 0.55                           & 20                             & 1                             & 0.3                          & 11                                      & 6                                       & 5                                       \\
B        & 0.64                           & 20                             & 1                             & 0.12                         & 12.8                                    & 11.4                                    & 1.4                                     \\
\textbf{C}        & \textbf{0.4}                            & 20                             & 1                             & -                            & -                                       & -                                       & -                                       \\
D        & 0.8                            & 20                             & 1                             & 0.13                         & 16                                      & 14.4                                    & 1.6                                     \\
E        & 0.9                            & 20                             & 1                             & 0.22                         & 18                                      & 14.6                                    & 3.4                                     \\
\textbf{F}        & 0.51                           & 20                             & 1                             & \textbf{-1.2}                         & -                                       & -                                       & -                                       \\ \hline
\end{tabular}
}
\vspace*{-2mm}
\end{table}

The second part of the problem is deciding when we treat a given case assuming that intervention fulfills the required conditions, i.e., $P_{uout} > \tau$ and $CATE_1 > 0$. We use the \emph{resource allocator} to tackle this part. 

The resource allocator monitors the availability of resources to allocate them efficiently. Allocating resources to $c_{id}$ raises another question: how long, i.e., treatment duration, the allocated resource is blocked to apply $T_{i=1}$. 

A simple way to define the treatment duration (hereafter $T_{dur}$) is to set it as a fixed value based on the domain knowledge. However, the variability of $T_{dur}$ might affect the net gain; therefore, we examine three different distributions for the $T_{dur}$, i.e., fixed, normal, and exponential. 

Finally, based on the domain knowledge that tells us how many resources are available to apply $T_{i=1}$, we keep an ordered list of the max gains for each running case $c_{id}$. Once we have an available resource, we allocate it to apply $T_{i=1}$ to $c_{id}$ with the max gain in our ordered list and block it for $T_{dur}$.

For example, in table \ref{tab:gainex}, suppose $res_1$ and $res_2$ are available. First, we allocate $res_1$ to $c_{id} = A$ and $res_2$ to $c_{id} =B$ and block them for $T_{dur}$. Then, $c_{id} = D$ enters; but, we can not treat it since there are no available resources. Accordingly, we keep $c_{id} = D$ and $c_{id} = E$ (that comes later) in our sorted list. We assume that cases keep coming to our system, implying that the sorted list may eventually be extended with other cases with positive gains. Whenever a resource becomes available, if a case in the sorted list has a positive gain, we allocate resources to the one with max gain. Moreover, our approach allocates resources to different cases simultaneously, and various instances update the available resources.



\vspace{-4mm}

\vspace{-1mm}
\section{Evaluation} \label{sec:evaluation}
\vspace{-1mm}

We conducted an evaluation to address the following research questions: 
    \begin{questions}
        \item To what extent the total gain depends on the number of available resources?\label{rq:rq1}
        \item To what extent the total gain depends on the variability of the $T_{dur}$?\label{rq:rq2} 
        \item When allocating resources to cases with higher gain versus cases with higher undesired outcome probability, what is the total gain?\label{rq:rq3}
        
    \end{questions}



\vspace{-1mm}
\subsection{Dataset}  \label{data}
\vspace{-1mm}
We evaluate our approach using one real-life event log, namely \emph{BPIC2017}\footnote{\url{https://doi.org/10.4121/uuid:5f3067df-f10b-45da-b98b-86ae4c7a310b}}, corresponding to a loan origination process. In this event log, each case corresponds to a loan application. Each application has an outcome. The desired one occurs when offering clients a loan, and clients accept and sign it. While the undesired one occurs when the bank cancels the application or the client rejects the offer. The log contains $31,413$  applications and $1,202,267$ events.




We used all possible attributes in the log as input to the predictive and causal models. Furthermore, we extracted other features, e.g., the number of offers, event number, and other temporal information, e.g., the hour of the day, day of the month, and month. We extracted prefixes of length less than or equal to the $90^{th}$ percentile of the case lengths in the log to avoid bias from long cases. We encoded the extracted prefixes using \emph{aggregate encoding} to convert them into a fixed-size feature vector (see Sect. 2.1). 

To obtain the best performance of either predictive or causal models, event log, i.e., a \emph{loan application process}, preprocessing is an essential step. In addition to the preprocessing given by \cite{teinemaa2019outcome}, we define the outcome of cases based on the end activity. We represent cases that end with \emph{“A\_Pending”} events as a positive outcome, where cases that have \emph{“A\_Denied”} or \emph{“A\_Cancelled”} events are adverse outcomes that need intervention. Then, we define the intervention that we could apply to minimize the unsuccessful loan applications based on the winner report of the BPIC challenge \cite{povalyaeva2017bpic}. They report that making more offers to clients increases the probability of having \emph{“A\_Pending”} as an end stat. Accordingly, we represent cases with only one offer to be treated where $T=1$. In contrast, cases with more than one offer should not be
treated, then $T =0.$


\vspace{-2mm}
\subsection{Experiment setup} \label{exp_setup}

We use an \emph{XGBoost}\footnote{\url{https://github.com/dmlc/xgboost}} model to estimate the probability of negative case outcomes, i.e., $P_{uout}$. XGBoost has shown good results on various classification problems \cite{fernandez2014we}, including outcome-oriented PPM~\cite{teinemaa2019outcome}. 
We use the following parameters to train the XGBoost model: learning rate of $0.2$, subsample of $0.89$, max tree depth of $14$, Colsample by tree  of $0.4$, and a min child weight of $3$. 

We use \emph{ORF} to estimate the $CATE$ as implemented in the \emph{EconMl}\footnote{\url{https://github.com/microsoft/EconML}} package. 
We use the following parameters: min leaf size of $50$, a max depth of $20$, a sub-sample ratio of $0.4$, and lambda regularization with parameter $0.01$.



\enlargethispage{-1\baselineskip}
The predictive and causal models follow the same workflow as any machine learning problem. We temporally split the log into three parts ($60\%$ - $20\%$ - $20\%$) to simulate real-life situations to tune and evaluate these models. Mainly, we arrange cases using their timestamps. We use the opening $80\%$ for training ($60 \%$) and tuning ($20 \%$), and the rest ($20 \%$) to evaluate model performance. Table~\ref{tab:config} shows the configurations of the proposed approach.  

We vary the $c_{uout}$ values to make them more significant than the $c_{T_1}$ value to give a meaningful result. We found that the higher $c_{uout}$ related to $c_{T_1}$, the more net gain. Accordingly, we applied the higher value of the $c_{uout}$ in our experiments with different treatment distributions and an empirically optimized threshold to answer our research questions. 




We assume that the estimated $CATE$ is accurate and, hence, allocating resources will decrease a case’s probability of a negative outcome. We compare our approach to a purely predictive baseline proposed in \cite{fahrenkrog2019fire}, where the interventions are triggered as soon as $P_{uout} > \tau$. In other words, we allocate resources to cases with the highest $P_{uout}$ instead of cases with max gain, and we consider the $CATE$ as the new gain we achieve from treating cases.


\begin{table}[]
\caption{Configurations of the proposed approach}
\label{tab:config}
\centering
\begin{tabular}{|l|l|l|l|l|}
\hline
\#$resources$        & $c_{uout}$      & $c_{T_1}$    & $\tau$  \hspace{3cm}  & $T\_{dur}$ (sec)\\ \hline
$1, 2, ... 10$  & $1,2,3,5,10,20$ & 1  & 
$0.5, 0.6, ... 0.9$  & Fixed = $60$ \\ &&&& Normal $\in$ \{$1, 60$\}\\ &&&& Exponential $\in$ \{$1, 60$\}
 \\
\hline 
\end{tabular}
\end{table}



\vspace{-1mm}
\subsection{Results} \label{results}
\begin{figure}[!h]
\vspace*{-4mm}
\centering
\begin{subfigure}{.5\textwidth}
  \centering
  \includegraphics[width=\linewidth, ]{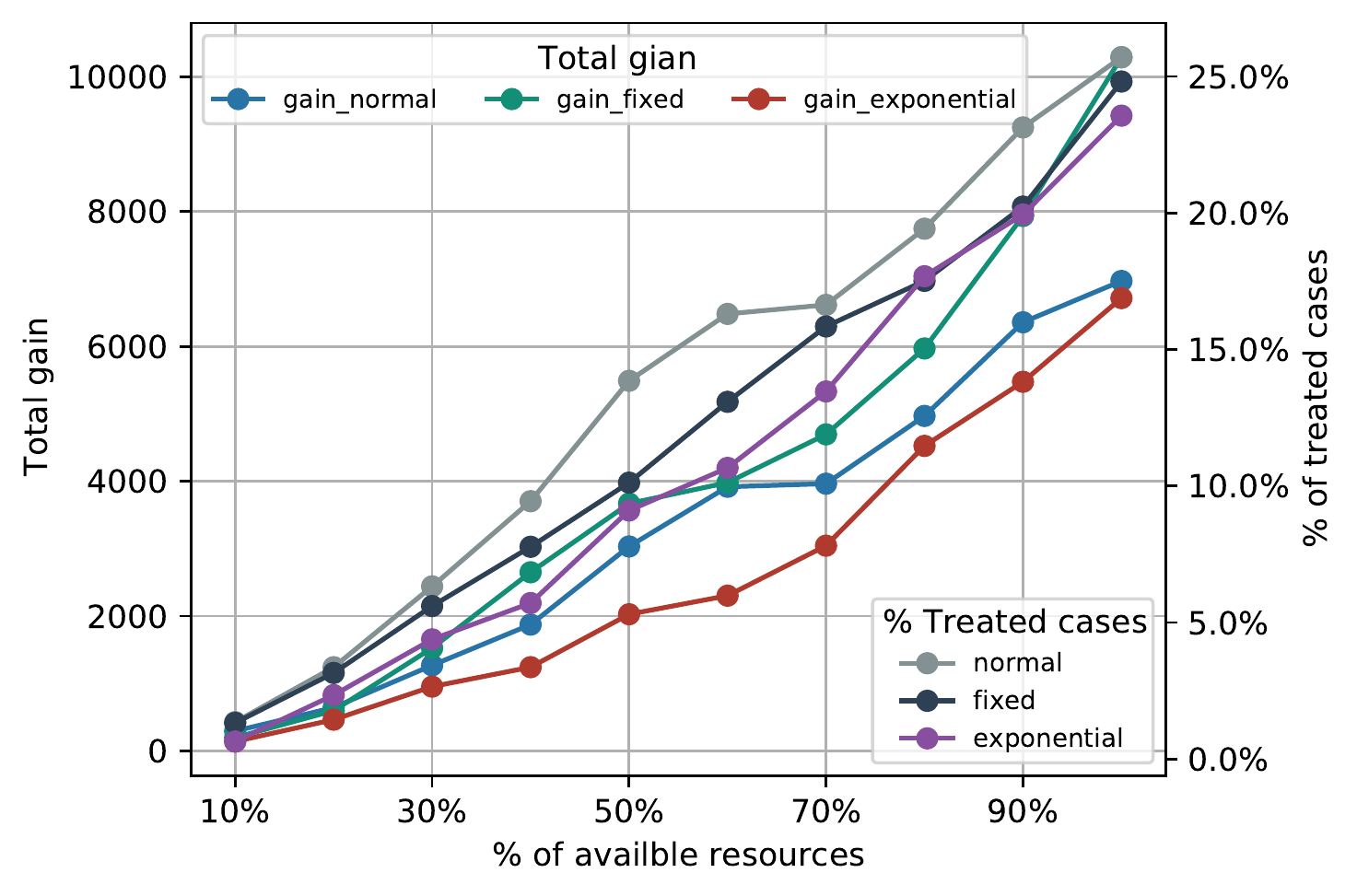}
  \vspace*{-2mm}
  \scalebox{0.7}{\parbox{\linewidth}{%
  \caption{RQ1 and RQ2}
  \label{fig:res_rq12}}}
\end{subfigure}%
\begin{subfigure}{.5\textwidth}
  \centering
  \includegraphics[width=\linewidth]{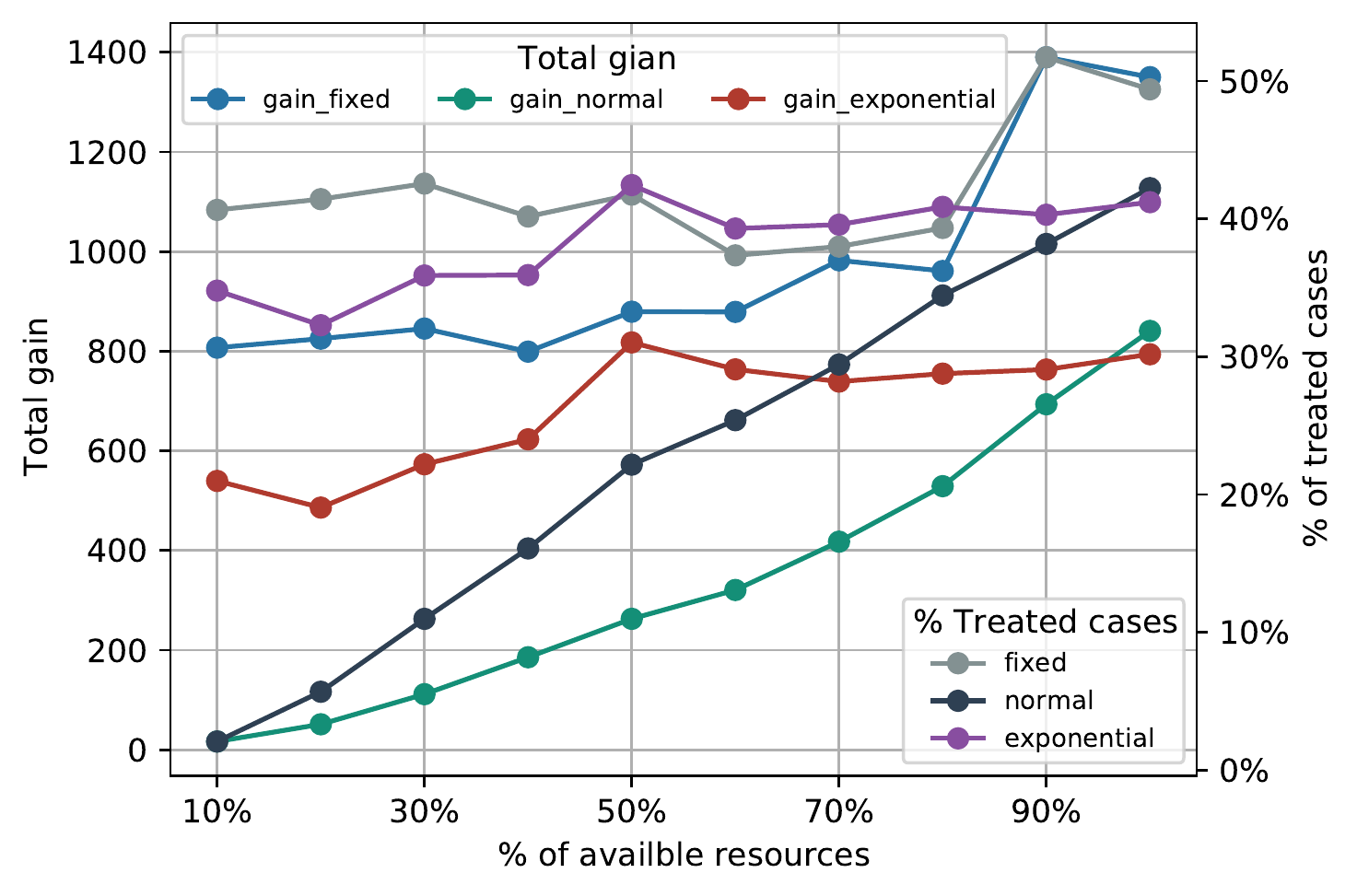}
  \vspace*{-2mm}
  \scalebox{0.7}{\parbox{\linewidth}{%
  \caption{RQ3}
  \label{fig:res_rq3}}}
\end{subfigure}
\vspace*{-2mm}
\caption{Total gain and \% of treated cases w.r.t \% available resources}
\vspace*{-2mm}
\label{fig:test}
\end{figure}

\vspace*{-1mm}
We present the results of our proposed approach by exploring the effects of available resources on the total gain and the percentage of treated cases, taking into account the variability of $T_{dur}$ (\ref{rq:rq1} and \ref{rq:rq2}).  
Fig \ref{fig:res_rq12} shows how the total gain and percentage of treated cases evolve as we increase the number of available resources (\ref{rq:rq1}). When the number of available resources increases, both metrics increase. Meanwhile, if the available resources reach above $50\%$, the total gain almost increases considerably.  For example, with fixed distribution in Fig \ref{fig:res_rq12}, when the percentage of available resources is above $50\%$, the total gain rises markedly compared to the situation where the rate of resources is below $50 \%$. That is because more cases are treated when more than half of the resources become available.


\vspace{-6mm}
Moving to \ref{rq:rq2}, we experiment with three $T_{dur}$ distributions, i.e., fixed, normal, and exponential.  Fig \ref{fig:res_rq12} shows that the fixed distribution gives more net gain because there is less variability in the distribution of resources among cases that need intervention than normal and exponential distributions where the level of variability decreases, respectively. Accordingly, the net gain highly depends on the variability of treatment duration.


To answer \ref{rq:rq3}, we allocate resources to cases with the highest $P_{uout}$ instead of cases with max gain. We consider the $CATE$ a new gain from treating cases (see Fig \ref{fig:res_rq3}).  Therefore, we need a threshold $\tau$ to determine whether or not to intervene depending on the $P_{uout}$. There are two approaches to set a threshold: first, and based on a given threshold, e.g., $\tau = 0.5$, if there are available resources and the undesired outcome above the given threshold, we trigger an intervention. The second is to use an empirical threshold proposed by \cite{fahrenkrog2019fire}, where authors compute an optimal threshold based on historical data. We varied the threshold as shown in table \ref{tab:config}. However, the results are different based on the $T_{dur}$ distribution. Where $\tau = 0.5$, the normal distribution gives more net gain than other thresholds. While $\tau = 0.6$, the exponential distribution delivers the higher net gain. Moreover, with $\tau = 0.7,$ the fixed distribution wins. The results of optimizing the threshold are available in the supplementary material\footnote{\url{https://zenodo.org/record/5538113\#.YVSdhSVRV8I}}, where we show how the total gain changes w.r.t different thresholds and different $T_{dur}$.

We observe that our approach consistently leads to higher net gain, under the same amount of consumed resources, than the purely predictive baseline. For example, under a fixed distribution, treating $25\%$ of cases with our approach (cf.~Fig \ref{fig:res_rq12}) leads to a net gain of $10000$, while in the predictive method (Fig \ref{fig:res_rq3}), treating twice more cases ($50\%$ of cases) yields a net gain of only $1400$. This suggests that the combination of causal inference with predictive modeling can enhance the efficiency of prescriptive process monitoring methods.


\vspace*{-3mm}
\subsection{Threats to Validity}
\vspace*{-2mm}

The evaluation comes with an external validity threat (lack of generalizability) due to its reliance on only one event log. The evaluation is preliminary and ought to be followed up with additional experiments using other datasets.

We simulated a scenario where we could trigger interventions at any time point. Also, we assume that the effect of this intervention will be to reduce the probability of adverse outcomes by the estimated $CATE$. There is a threat to ecological validity because the $CATE$ might not reflect the actual treatment effect due to unobserved confounders.

The evaluation is limited to one feature encoding method and one machine learning algorithm. Experimenting with other encodings and algorithms is a direction for future work.

\vspace{-3mm}
\section{Conclusion} \label{sec:conclusion}
\vspace{-2mm}
We introduced a prescriptive monitoring approach that triggers interventions in ongoing cases of a process to maximize a net gain function under limited resources. The approach combines a predictive model to identify cases that are likely to end in a negative outcome (and hence create a cost) with a causal model to determine which cases would most benefit from the intervention in their current state. These two models are embedded into an allocation procedure that allocates resources to case interventions based on their estimated net gain. 
A preliminary evaluation suggests that it treats fewer cases and allocates resources more effectively than a baseline method that relies only on a predictive model.

In the proposed approach, an intervention is triggered whenever the estimated net gain of treating this case is maximal, relative to other cases. Under some circumstances, this may lead to treating a case at a suboptimal time. For example, in a loan origination process, calling a customer two days after sending an offer may be more effective than doing so just one day after the offer. The expected gain is not just depending on utilizing the intervention. It rather depends on the time we trigger the intervention. Accordingly, If we decide to wait until the state of the cases changes and do not intervene, it will reduce the uncertainty and probably achieving more gain. Our approach would trigger the intervention “call customer” one day after the offer if the expected benefit is positive and there is no other case with a higher net gain. An alternative approach would be to allocate resources based on the estimated net gain of a case intervention at the current time and the expected gain of intervening in the same case at a future time. An avenue for future work is to combine the proposed approach with a method that optimizes the time point when an intervention is triggered in a case. 
A related avenue for future work is to consider constraints on the moment when interventions can be triggered on a case. For example, calling a customer to follow up on a loan offer does not make sense if the loan offer has been canceled or the customer has not yet received a loan offer. 

Another limitation of the proposed approach is that it assumes that there is a single type of intervention. In reality, there may be multiple types of interventions (e.g., \ call the customer, send a second loan offer). Another future work direction is to handle multiple types of interventions.

\medskip\noindent\textbf{Reproducibility.} The implementation and source code of our approach can be found at \url{https://github.com/mshoush/PrescriptiveProcessMonitoring}.

\vspace*{-5mm}
\bibliographystyle{splncs04}
\bibliography{References}
\vspace{-5mm}

\end{document}